\documentclass[runningheads]{llncs}
\usepackage{float}
\usepackage{graphicx}
\usepackage{cite}
\usepackage{amsmath,amssymb,amsfonts}
\usepackage{stmaryrd}
\usepackage{algorithmic}
\usepackage{algorithm}
\usepackage[english]{babel}
\usepackage{graphicx}
\usepackage{textcomp}
\usepackage{multirow}
\usepackage{multicol}
\usepackage{bm}
\usepackage{mathtools}
\usepackage[makeroom]{cancel}
\usepackage[dvipsnames]{xcolor}
\usepackage{booktabs}       
\usepackage{mathfont}
\usepackage{verbatim}
\usepackage{caption}
\usepackage{subcaption}
\newcommand{\IfElse}[3]{
   \textbf{if}\ #1\ \textbf{then}\ #2\ \textbf{else}\ #3}

\newtheorem{lem}{Lemma}[subsection]

\DeclareMathOperator*{\argmin}{arg\,min}
\newcommand{\defeq}{\vcentcolon=}
\newcommand\norm[1]{\left\lVert#1\right\rVert}
\newcommand{\innerproduct}[2]{\langle #1, #2 \rangle}
\def\*#1{\mathbf{#1}}
\def\BibTeX{{\rm B\kern-.05em{\sc i\kern-.025em b}\kern-.08em
    T\kern-.1667em\lower.7ex\hbox{E}\kern-.125emX}}
\newcommand{\expnum}[2]{{#1}{\text{e}}{#2}}
\newcommand{\expb}[2]{{#1}{e}{#2}}
\begin{document}
\title{Learned Alternating Minimization Algorithm for Dual-domain Sparse-View CT Reconstruction\thanks{This work was supported in part by National Science Foundation under grants DMS-1925263, DMS-2152960 and DMS-2152961 and US National Institutes of Health grants R01HL151561, R01CA237267, R01EB032716 and R01EB031885.}}
\titlerunning{LAMA}
%
\author{Chi Ding\inst{1}, Qingchao Zhang\inst{1}, Ge Wang\inst{2}, Xiaojing Ye\inst{3}
\and Yunmei Chen\inst{1}
}
\institute{University of Florida, Gainesville FL 32611, USA\\
\email{\{ding.chi, qingchaozhang, yun\}@ufl.edu}\\
\and Rensselaer Polytechnic Institute, Troy NY 12180, USA\\
\email{wangg6@rpi.edu}
\and Georgia State University, Atlanta GA 30302, USA\\
\email{xye@gsu.edu}\\
}

%
\authorrunning{Ding et al.}
%
%
\maketitle              
\begin{abstract}
We propose a novel Learned Alternating Minimization Algorithm (LAMA) for dual-domain sparse-view CT image reconstruction. LAMA is naturally induced by a variational model for CT reconstruction with learnable nonsmooth nonconvex regularizers, which are parameterized as composite functions of deep networks in both image and sinogram domains. To minimize the objective of the model, we incorporate the smoothing technique and residual learning architecture into the design of LAMA. We show that LAMA substantially reduces network complexity, improves memory efficiency and reconstruction accuracy, and is provably convergent for reliable reconstructions. 
Extensive numerical experiments demonstrate that LAMA outperforms existing methods by a wide margin on multiple benchmark CT datasets.  
%

\end{abstract}
\section{Introduction}
Sparse-view Computed Tomography (CT) is an important class of low-dose CT techniques for fast imaging with reduced X-ray radiation dose. 
Due to the significant undersampling of sinogram data, the sparse-view CT reconstruction problem is severely ill-posed.
As such, applying the standard filtered-back-projection (FBP) algorithm, \cite{kak_slaney_2001} to sparse-view CT data results in significant severe artifacts in the reconstructed images, which are unreliable for clinical use. 
%
In recent decades, variational methods have become a major class of mathematical approaches that model reconstruction as a minimization problem. The objective function of the minimization problem consists of a penalty term that measures the discrepancy between the reconstructed image and the given data and a regularization term that enforces prior knowledge or regularity of the image. Then an optimization method is applied to solve for the minimizer, which is the reconstructed image of the problem. The regularization in existing variational methods is often chosen as relatively simple functions, such as total variation (TV) \cite{RUDIN1992259,accFV,Kim_2016}, which is proven useful in many instances but still far from satisfaction in most real-world image reconstruction applications due to their limitations in capturing fine structures of images. 
Hence, it remains a very active research area in developing more accurate and effective methods for high-quality sparse-view CT reconstruction in medical imaging.

Deep learning (DL) has emerged in recent years as a powerful tool for image reconstruction. Deep learning parameterizes the functions of interests, such as the mapping from incomplete and/or noisy data to reconstructed images, as deep neural networks. The parameters of the networks are learned by minimizing some loss functional that measures the mapping quality based on a sufficient amount of data samples. The use of training samples enables DL to learn more enriched features, and therefore, DL has shown tremendous success in various tasks in image reconstruction. In particular, DL has been used for medical image reconstruction applications \cite{DDNet, DuDoTrans, sinSyn, DRONE, FBPConvNet,LEARN,GamReg,RED-CNN}, and experiments show that these methods often significantly outperform traditional variational methods. 

DL-based methods for CT reconstruction have also evolved fast in the past few years. One of the most successful DL-based approaches is known as unrolling \cite{ISTANet,unrolling,LEARN,learnPP,LDA}. Unrolling methods mimic some traditional optimization schemes (such as proximal gradient descent) designed for variational methods to build the network structure but replace the term corresponding to the handcrafted regularization in the original variational model by deep networks. Most existing DL-based CT reconstruction methods use deep networks to extract features of the image or the sinogram \cite{LEARN,GamReg, DDNet, FBPConvNet,RED-CNN, sinSyn, TransItr, DDPNet, ldct}. More recently, dual-domain methods \cite{learnPP, DRONE, DuDoTrans, DDPNet} emerged and can further improve reconstruction quality by leveraging complementary information from both the image and sinogram domains. 
Despite the substantial improvements in reconstruction quality over traditional variational methods, there are concerns with these DL-based methods due to their lack of theoretical interpretation and practical robustness. In particular, these methods tend to be memory inefficient and prone to overfitting. 
One major reason is that these methods only superficially mimic some known optimization schemes but lose all convergence and stability guarantees.
%

Recently, a new class of DL-based methods known as learnable descent algorithm (LDA) \cite{LDA,ldct,wanyuMiccai} have been developed for image reconstruction. These methods start from a variational model where the regularization can be parameterized as a deep network whose parameters can be learned. The objective function is potentially nonconvex and nonsmooth due to such parameterization. Then LDA aims to design an efficient and convergent scheme to minimize the objective function. This optimization scheme induces a highly structured deep network whose parameters are completely inherited from the learnable regularization and trained adaptively using data while retaining all convergence properties. The present work follows this approach to develop a dual-domain sparse-view CT reconstruction method. Specifically, we consider learnable regularizations for image and sinogram as composite objectives, where they unroll parallel subnetworks and extract complementary information from both domains. Unlike the existing LDA, we will design a novel adaptive scheme by modifying the alternating minimization methods \cite{ProxSARAH,PALM,inertialProx,stochProx,inexBlock} and incorporating the residual learning architecture to improve image quality and training efficiency. 

\section{Learnable Variational Model}

We formulate the dual-domain reconstruction model as the following two-block minimization problem: 
\begin{equation}
    \label{eq:OrgPhi}
\argmin_{\*x,\*z}\,\Phi(\*x,\*z;\*s,\Theta)\defeq \frac 1 2 \norm{\*{Ax}-\*z}^2 + \frac \lambda 2 \norm{\*P_{0}\*z - \*s}^2+R(\*x;\theta_1)+Q(\*z;\theta_2),
\end{equation}
where $(\*x,\*z)$ are the image and sinogram to be reconstructed and $\*s$ is the sparse-view sinogram. The first two terms in \eqref{eq:OrgPhi} are the data fidelity and consistency, where $\*A$ and $\*P_0\zbf$ represent the Radon transform and the sparse-view sinogram, respectively, and $\norm{\cdot} \equiv \norm{\cdot}_2$. The last two terms represent the regularizations, which are defined as the $l_{2,1}$ norm of the learnable convolutional feature extraction mappings in \eqref{eq:21norm}. If this mapping is the gradient operator, then the regularization reduces to total variation that has been widely used as a hand-crafted regularizer in image reconstruction. On the other hand, the proposed regularizers are strict generalizations and capable to learn in more adapted domains where the reconstructed image and sinogram become sparse: 
\begin{subequations}
\label{eq:21norm}
\begin{eqnarray}
    R(\*x;\theta_1)=\norm{\*g^R(\*x, \theta_1)}_{2,1}&\defeq\sum_{i=1}^{m_R} \norm{\*g^R_i(\*x, \theta_1)}, \\
    Q(\*z;\theta_2)=\norm{\*g^Q(\*z, \theta_2)}_{2,1}&\defeq\sum_{j=1}^{m_Q} \norm{\*g^Q_j(\*z, \theta_2)},
\end{eqnarray}
\end{subequations}
where $\theta_1, \ \theta_2$ are learnable parameters. We use $\*g^r(\cdot)\in\mathbb{R}^{m_r\times d_r}$ to present $\*g^R(\*x, \theta_1)$ and $\*g^Q(\*z, \theta_2)$, i.e. $r$ can be $R$ or $Q$. The $d_r$ is the depth and 
$\sqrt{m_r}\times\sqrt{m_r}$ is the spacial dimension. Note $\*g^r_i(\cdot) \in \mathbb{R}^{d_r}$ is the vector at position $i$ across all channels. The feature extractor $\*g^r(\cdot)$ is a  CNN consisting of several convolutional operators separated by  the smoothed ReLU activation function as follows:
\begin{equation}
    \label{eq:g_arch}
    \*g^r(\*y) = \*w_l^r * a \cdots a(\*w_2^r * a(\*w_1^r * \*y)),
\end{equation}
where $\{\*w_i^r\}_{i=1}^l$ denote convolution parameters with $d_r$ kernels. Kernel sizes are $(3,3)$ and $(3,15)$ for the image and sinogram networks, respectively. $a(\cdot)$ denotes smoothed ReLU activation function, which can be found in \cite{LDA}.

\section{A Learned Alternating Minimization Algorithm}
This section formally introduces the Learned Alternating Minimization Algorithm (LAMA) to solve the nonconvex and nonsmooth minimization model \eqref{eq:OrgPhi}. LAMA incorporates the residue learning structure \cite{ResNet} 
to improve the practical learning performance by avoiding gradient vanishing in the training process 
with convergence guarantees. The algorithm consists of three stages, as follows:

The first stage of LAMA aims to reduce the nonconvex and nonsmooth problem in \eqref{eq:OrgPhi} to a nonconvex smooth optimization problem by using an appropriate smoothing procedure 
\begin{equation}
    \label{eq:smoothedR}
    r_\varepsilon(\*y) = \sum_{i\in I^r_0}\frac{1}{2\varepsilon}\norm{\*g^r_i(\*y)}^2 + \sum_{i\in I_1^{r}}\left( \norm{\*g^r_i(\*y)}-\frac \varepsilon 2\right), \quad \ybf \in Y=:m_r \times d_r,
\end{equation}
where $(r,\*y)$ represents either $(R, \*x)$ or $(Q, \*z)$ and 
\begin{equation}
    I_0^r=\{i\in \left[m_r\right]\,\vert\norm{\*g^r_i(\*y)}\leq \varepsilon\},\quad I_1^r = \left[m_r\right] \setminus I_0^r.
\end{equation}
Note that the non-smoothness of the objective function \eqref{eq:OrgPhi} originates from the non-differentiability of the $l_{2,1}$ norm at the origin. To handle the non-smoothness, we utilize Nesterov's smoothing technique \cite{nesterov_2004} as previously applied in \cite{LDA}. The smoothed regularizations take the form of the Huber function, effectively removing the non-smoothness aspects of the problem.

The second stage solves the smoothed nonconvex problem with the fixed smoothing factor $\varepsilon=\varepsilon_k$, i.e. 
\begin{equation}
 \label{SM}   
\min_{\*x, \*z}\{\Phi_{\varepsilon}(\*x, \*z):=f(\*x,\*z)+R_{\varepsilon}(\*x)+Q_{\varepsilon}(\*z)\}.
\end{equation}
where $f(\*x,\*z)$ denotes the first two data fitting terms from \eqref{eq:OrgPhi}. In light of the substantial improvement in practical performance by ResNet \cite{ResNet},  we propose an inexact  proximal alternating linearized minimization algorithm (PALM) \cite{PALM} for solving \eqref{SM}.  With $\varepsilon = \varepsilon_k >0$, the scheme of PALM \cite{PALM} is 
\begin{equation}
    \*b_{k+1} =\*z_k-\alpha_k \nabla_{\*z} f(\*x_k,\*z_k), \quad
     \*u_{k+1}^{\*z} =\argmin _{\*u}  \frac{1}{2\alpha_k}\norm{\*u-\*b_{k+1}}^2+Q_{\varepsilon_k}(\*u),
\end{equation}

\begin{equation}
   \*c_{k+1} =\*x_k-\beta_k \nabla_{\*x} f(\*x_k,\*u^{\*z}_{k+1}), \quad
 \*u_{k+1}^{\*x} = \argmin_{\*u} \frac{1}{2\beta_k}\norm{\*u-\*c_{k+1}}^2+R_{\varepsilon_k}(\*u),
\end{equation}
where $\alpha_k$ and $\beta_k$ are step sizes. Since the proximal point $\*u^{\*x}_{k+1}$ and $\*u^{\*z}_{k+1}$ are are difficult to compute, we approximate $Q_{\varepsilon_k}(\*u)$ and $R_{\varepsilon_k}(\*u)$  by their linear approximations at $\*b_{k+1}$ and $\*c_{k+1}$, i.e. $Q_{\varepsilon_k}(\*b_{k+1})+ \innerproduct{\nabla Q_{\varepsilon_k}(\*b_{k+1})}{ \*y-\*b_{k+1}}$ and $R_{\varepsilon_k}(\*c_{k+1})+ \innerproduct{\nabla R_{\varepsilon_k}(\*c_{k+1})}{ \*u-\*c_{k+1}}$, together with the proximal terms $\frac{1}{2p_k} \norm{\*u-\*b_{k+1}}^2$ and $\frac{1}{2q_k} \norm{\*u-\*c_{k+1}}^2$. Then by a simple computation, $\*u^{\*x}_{k+1}$ and $\*u^{\*z}_{k+1}$ are now determined by the following formulas
\begin{equation}
    \label{eq:u}
        \*u^{\*z}_{k+1} = \*b_{k+1}-\hat{\alpha}_k\nabla Q_{\varepsilon_k}(\*b_{k+1}),\quad
        \*u^{\*x}_{k+1} = \*c_{k+1}-\hat\beta_k\nabla R_{\varepsilon_k}(\*c_{k+1}),
\end{equation}
where $\hat\alpha_k = \frac{\alpha_k p_k}{\alpha_k + p_k}$, $\hat\beta_k = \frac{\beta_k q_k}{\beta_k + q_k}$. In deep learning approach, the step sizes $\alpha_k$, $\hat\alpha_k$, $\beta_k$ and $\hat\beta_k$ can also be learned. Note that the convergence of the sequence $\{(\*u^{\*z}_{k+1}, \*u^{\*x}_{k+1})\}$ is not guaranteed. 
We proposed that if $(\*u^{\*z}_{k+1}, \*u^{\*x}_{k+1})$ satisfy the following \textbf{Energy Descent Conditions} (EDC):
\begin{subequations}
\label{eq:uvcond}
    \begin{align}
    \label{eq:uvconda}
    \Phi_{\varepsilon_k}(\*u^{\*x}_{k+1},\*u^{\*z}_{k+1})-\Phi_{\varepsilon_k}(\*x_k,\*z_k) &\leq -\eta \left(\norm{\*u^{\*x}_{k+1}-\*x_k}^2+\norm{\*u^{\*z}_{k+1}-\*z_k}^2\right),\\
    \label{eq:uvcondb}
       \norm{\nabla \Phi_{\varepsilon_k}(\*x_{k},\*z_{k})}&\leq \frac{1}{\eta} \left(\norm{\*u^{\*x}_{k+1}-\*x_k}+\norm{\*u^{\*z}_{k+1}-\*z_k}\right),
    \end{align}
\end{subequations}
for some $\eta>0$, we accept  $\*x_{k+1}=\*u_{k+1}^{\*x},\quad \*z_{k+1}=\*u_{k+1}^{\*z} $. 
If  one of \eqref{eq:uvconda} and \eqref{eq:uvcondb} is violated, we compute $(\*v^{\*z}_{k+1},\*v^{\*x}_{k+1})$ by the standard Block Coordinate Descent (BCD) with a simple line-search strategy to safeguard convergence: Let $\bar \alpha,\bar \beta$ be positive numbers in $(0,1)$
compute
\begin{align} \label{v1}
  &  \*v^{\*z}_{k+1}
  =  \*z_k-\bar \alpha\left( \nabla_{\*z} f(\*x_k,\*z_k)+\nabla Q_{\varepsilon_k}(\*z_k)\right),\\\label{v2}
 &   \*v_{k+1}^{\*x}
     =\*x_k-\bar \beta\left( \nabla_{\*x} f(\*x_k,\*v_{k+1}^{\*z})+\nabla R_{\varepsilon_k}(\*x_k)\right).
\end{align}
Set
$\*x_{k+1}=\*v_{k+1}^{\*x},\quad \*z_{k+1}=\*v_{k+1}^{\*z}$, if for some $\delta\in (0,1)$, the following holds:
\begin{equation}
\label{v-condition-6}
\Phi_{\varepsilon}(\*v_{k+1}^{\*x},\*v_{k+1}^{\*z})-\Phi_{\varepsilon}(\*x_k,\*z_k)
\leq -\delta (\norm{\*v_{k+1}^{\*x}-\*x_k}^2+\norm{\*v_{k+1}^{\*z}-\*z_k}^2).
\end{equation}
Otherwise we reduce $(\bar \alpha,\bar \beta)\leftarrow \rho (\bar \alpha,\bar \beta)$ where $0<\rho<1$, and recompute $\*v^{\*x}_{k+1},\*v^{\*z}_{k+1}$ until the condition (\ref{v-condition-6}) holds. 

The third stage checks if  $\|\nabla \Phi_{\varepsilon}\|$  has been reduced enough to  perform the second stage with a reduced  smoothing factor $\varepsilon$.  By gradually decreasing $\varepsilon$, we obtain a subsequence of the iterates that converges to a Clarke stationary point of the original nonconvex and nonsmooth problem. The algorithm is given below.

\begin{algorithm}[htb]
\caption{The Linearized Alternating Minimization Algorithm (LAMA)}
\textbf{Input:} Initializations: $\*x_0$, $\*z_0$, $\delta$, $\eta$, $\rho$, $\gamma$, $\varepsilon_0$, $\sigma$, $\lambda$
\begin{algorithmic}[1]
    \FOR{$k=0,\,1,\,2,\,...$}
        \STATE $\*b_{k+1} =\*z_k-\alpha_k \nabla_{\*z} f(\*x_k,\*z_k),\,\*u^{\*z}_{k+1} = \*b_{k+1}-\hat{\alpha}_k\nabla Q_{\varepsilon_k}(\*b_{k+1})$\label{alg:begin}
        \STATE{$\*c_{k+1} =\*x_k-\beta_k \nabla_{\*x }f(\*x_k,\*u^{\*z}_{k+1}),\,\*u^{\*x}_{k+1} = \*c_{k+1}-\hat\beta_k\nabla R_{\varepsilon_k}(\*c_{k+1})$} \label{alg:c}
        \IF{\eqref{eq:uvcond} holds}
            \STATE{$(\*x_{k+1}, \*z_{k+1})\leftarrow (\*u^{\*x}_{k+1}, \*u^{\*z}_{k+1})$}
        \ELSE
            \STATE{$\*v^{\*z}_{k+1}=\*z_k-\bar\alpha\left[ \nabla_{\*z}f(\*x_k,\*z_k)+ \nabla Q_{\varepsilon_k}(\*z_{k})\right]$}\label{alg:v}
            \STATE{$\*v^{\*x}_{k+1}=\*x_k-\bar\beta \left[\nabla_{\*x} f(\*x_k,\*v^{\*z}_{k+1})+\nabla  R_{\varepsilon_k}(\*x_k)\right]$}
            \STATE{\IfElse{\eqref{v-condition-6}}{$(\*x_{k+1}, \*z_{k+1})\leftarrow (\*v^{\*x}_{k+1}, \*v^{\*z}_{k+1})$}{$(\bar\beta, \bar\alpha) \leftarrow \rho (\bar\beta, \bar\alpha)$ and \textbf{go to} \ref{alg:v}}}
        \ENDIF
        \STATE{\IfElse{$\norm{\nabla \Phi_{\varepsilon_k}(\*x_{k+1}, \*z_{k+1})}<\sigma\gamma\varepsilon_k$}{$\varepsilon_{k+1}=\gamma\varepsilon_k$}{$\varepsilon_{k+1}=\varepsilon_k$}}
    \ENDFOR
    \RETURN $\*x_{k+1}$
\end{algorithmic}
\label{alg:LAMA}
\end{algorithm}

\section{Network Architecture}
\begin{figure}[htb]
    \centering
    \includegraphics[scale=0.43]{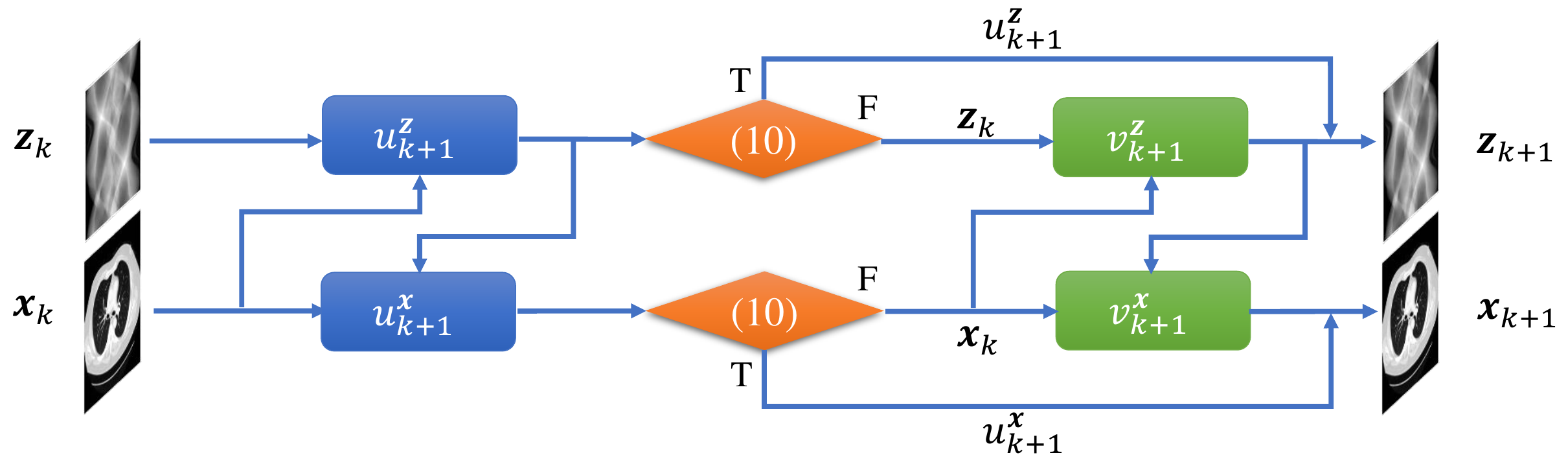}
    \caption{Schemetic illustration of one phase in LAMA, where \eqref{eq:uvcond} stands for the EDC.}
    \label{fig:flowchart}
\end{figure}

The architecture of the proposed multi-phase neural networks follows LAMA exactly. Hence we also use LAMA to denote the networks as each phase corresponds to each iteration in Algorithm \ref{alg:LAMA}. The networks inherit all the convergence properties of LAMA such that the solution is stabilized. Moreover, the algorithm effectively leverages complementary information through the inter-domain connections shown in Fig.~\ref{fig:flowchart} to accurately estimate the missing data. The network is also memory efficient due to parameter sharing across all phases. 

\section{Convergence Analysis}\label{sec:convergenceproof}
Since we deal with a nonconvex and nonsmooth optimization problem, we first need to introduce the following definitions based on the generalized derivatives.
\begin{definition} \label{def-clark-gen} (Clarke subdifferential). Suppose that $f:\mathbb{R}^n\times\mathbb{R}^m\rightarrow(-\infty,\infty]$ is locally Lipschitz. The Clarke subdifferential of $f$ at $(\xbf,\zbf)$ is defined as
    \begin{align*}
       & \partial^c f(\xbf,\zbf)\defeq\{(w_1,w_2)\in\mathbb{R}^n\times \mathbb{R}^m\vert \innerproduct{w_1}{v_1}+\innerproduct{w_2}{v_2}\\
       &\leq \limsup_{(z_1,z_2)\rightarrow(\xbf,\zbf),\,t\rightarrow0_+}\frac{f(z_1+tv_1,z_2+tv_2)-f(z_1,z_2)}{t},  \quad\forall (v_1,v_2)\in \mathbb{R}^n\times \mathbb{R}^m\}.
    \end{align*}where $\innerproduct{w_1}{v_1}$ stands for the inner product in $\mathbb{R}^n$ and similarly for $\innerproduct{w_2}{v_2}$.
\end{definition}

\begin{definition}
    (Clarke stationary point) For a locally Lipschitz function $f$ defined as in Def \ref{def-clark-gen}, a point $X=(\xbf, \zbf)\in \mathbb R^n\times \mathbb R^m$ is called a Clarke stationary point of $f$, if $0\in \partial f(X)$. 
\end{definition}

We can have the following convergence result. All proofs are given in the supplementary material. 
\begin{theorem} \label{thm}
    Let $\{Y_k=(\xbf_k, \zbf_k)\}$ be the sequence generated by the algorithm with arbitrary initial condition $Y_0=(\xbf_0, \zbf_0)$, arbitrary $\varepsilon_0>0$ and $\varepsilon_{tol}=0$. Let
    $\{\tilde Y_l\}
    =:(\xbf_{k_l+1}, \zbf_{k_l+1})\}$ be the subsequence, where the reduction criterion in the algorithm is met for $k=k_l$ and $l=1,2, ...$. Then $\{\tilde Y_l\}$ has at least one accumulation point, and each accumulation point is a Clarke stationary point. 
\end{theorem}
\section{Experiments and Results}
%
\subsection{Initialization Network}
The initialization $(\*x_0,\*z_0)$ is obtained by passing the sparse-view sinogram $\*s$ defined in \eqref{eq:OrgPhi} through a CNN consisting of five residual blocks. Each block has four convolutions with 48 channels and kernel size $(3,3)$, which are separated by ReLU. We train the CNN for 200 epochs using MSE, then use it to synthesize full-view sinograms $\*z_0$ from $\*s$. The initial image $\*x_0$ is  generated by applying FBP to $\*z_0$. The resulting image-sinogram pairs are then provided as inputs to LAMA for the final reconstruction procedure. Note that the memory size of our method in Table~\ref{tab:result} includes the parameters of the initialization network.
\begin{table}[htb]
    \centering
    \caption{Comparison of LAMA and existing methods on CT data with 64 and 128 views. 
    }
    \scalebox{0.77}{
    \begin{tabular}{|c|c|c|c|c|c|c|c|c|}
    \hline
    Data & Metric & Views & FBP\cite{kak_slaney_2001} & DDNet\cite{DDNet} & LDA\cite{LDA} & DuDoTrans\cite{DuDoTrans} & Learn++\cite{learnPP} & LAMA (Ours) \\\hline
    \multirow{4}{*}{Mayo}     & \multirow{2}{*}{PSNR} & 64 & $27.17 \pm 1.11$ & $35.70 \pm 1.50$ & $37.16 \pm 1.33 $ & $37.90 \pm 1.44$ & $43.02 \pm 2.08$ & $\*{44.58 \pm 1.15}$ \\
        &  & 128 & $33.28 \pm 0.85$ & $42.73 \pm 1.08$ & $43.00 \pm 0.91$ & $43.48 \pm 1.04$ & $49.77 \pm 0.96$ & $\*{50.01\pm0.69}$\\\cline{2-9}
        & \multirow{2}{*}{SSIM} & 64 & $0.596 \pm \expnum{9}{-4}$ & $0.923 \pm \expnum{4}{-5}$ & $0.932 \pm  \expnum{1}{-4}$ & $0.952\pm\expnum{1.0}{-4}$ & $0.980\pm\expnum{3}{-5}$ & $\*{0.986\pm\expb{7}{-6}}$\\
        & & 128 & $0.759 \pm \expnum{1}{-3}$ & $0.974 \pm \expnum{4}{-5}$ & $0.976\pm\expnum{2}{-5}$ & $0.985 \pm \expnum{1}{-5}$ & $0.995 \pm \expnum{1}{-6}$ & $\*{0.995\pm\expb{6}{-7}}$\\\hline
    \multirow{4}{*}{NBIA}     & \multirow{2}{*}{PSNR} & 64 & $25.72 \pm 1.93$ & $35.59 \pm 2.76$ & $34.31 \pm 2.20 $ & $35.53 \pm 2.63$ & $38.53 \pm 3.41$ & $\*{41.40 \pm 3.54}$ \\
        &  & 128 & $31.86 \pm 1.27$ & $40.23 \pm 1.98$ & $40.26 \pm 2.57$ & $40.67 \pm 2.84$ & $43.35 \pm 4.02$ & $\*{45.20\pm4.23}$\\\cline{2-9}
        & \multirow{2}{*}{SSIM} & 64 & $0.592 \pm \expnum{2}{-3}$ & $0.920 \pm \expnum{3}{-4}$ & $0.896 \pm  \expnum{4}{-4}$ & $0.938\pm\expnum{2}{-4}$ & $0.956\pm\expnum{2}{-4}$ & $\*{0.976\pm\expb{8}{-5}}$\\
        & & 128 & $0.743 \pm \expnum{2}{-3}$ & $0.961 \pm \expnum{1}{-4}$ & $0.963\pm\expnum{1}{-4}$ & $0.976 \pm \expnum{6}{-5}$ & $0.983 \pm \expnum{5}{-5}$ & $\*{0.988\pm\expb{3}{-5}}$\\\hline
        N/A & param. & N/A & N/A & $\expnum{6}{5}$ & $\*{\expb{6}{4}}$ & $\expnum{8}{6}$ & $\expnum{6}{6}$ & $\expnum{3}{5}$\\\hline
    \end{tabular}}
    \label{tab:result}
\end{table}
\subsection{Experiment Setup}\label{setup}
Our algorithm is evaluated on the ``\textit{2016 NIH-AAPM-Mayo Clinic Low-Dose CT Grand Challenge}" and the National Biomedical Imaging Archive (NBIA) datasets. We randomly select 500 and 200 image-sinogram pairs from AAPM-Mayo and NBIA, respectively, with 80\% for training and 20\% for testing. We evaluate algorithms using the peak signal-to-noise ratio (PSNR), structural similarity (SSIM), and the number of network parameters. The sinograms have 512 detector elements, each with 1024 evenly distributed projection views. The sinograms are downsampled into 64 or 128 views while the image size is $256\times256$, and we simulate projections and back-projections in fan-beam geometry using distance-driven algorithms \cite{distDriven,distDriven3} implemented in a PyTorch-based library CTLIB \cite{xia2021magic}. 
Given $N$ training data pairs $\{( \*s^{(i)}, \hat{\*x}^{(i)})\}_{i=1}^N$, the loss function for training the regularization networks is defined as:
\begin{equation}
    \mathcal{L}(\Theta) = \frac 1 N\sum_{i=1}^N \norm{\*x_{k+1}^{(i)}-\hat{\*x}^{(i)}}^2 + \norm{\*z_{k+1}^{(i)}-\*A\hat{\*x}^{(i)}}^2 + \mu\left(1-\text{SSIM}\big({\*x}_{k+1}^{(i)}, \hat{\*x}^{(i)}\big)\right),
\end{equation}where $\mu$ is the weight for SSIM loss set as $0.01$ for all experiments, $\hat{\*x}^{(i)}$ is ground truth image, and final reconstructions are $(\*x_{k+1}^{(i)}, \*z_{k+1}^{(i)}) \defeq \text{LAMA}(\*x_0^{(i)},\*z_0^{(i)})$.

We use the Adam optimizer with learning rates of 1e-4 and 6e-5 for the image and sinogram networks, respectively, and train them with a warm-up approach. The training starts with three phases for 300 epochs, then adding two phases for 200 epochs each time until the number of phases reaches 15. The algorithm is implemented in Python using the PyTorch framework and is available on GitHub. Our experiments were run on a Linux server with an NVIDIA A100 Tensor Core GPU. 

\begin{figure}[htb]
    \centering
    \includegraphics[scale=0.6]{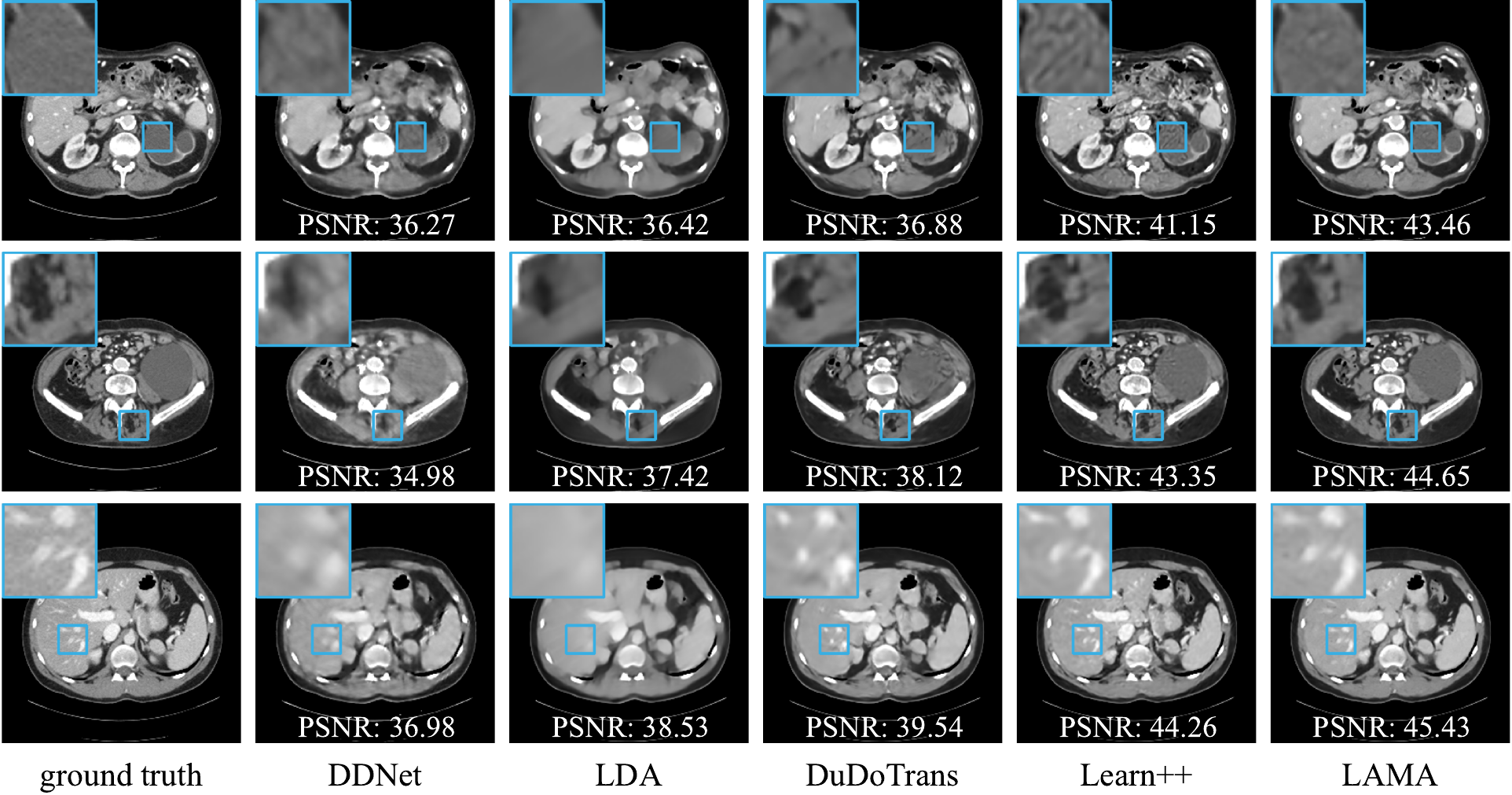}
    \caption{Visual comparison for AAPM-Mayo dataset using 64-view sinograms.}
    \label{fig:results}
\end{figure}
\subsection{Numerical and Visual Results}
\begin{figure}
    \centering
    \includegraphics[scale=0.35]{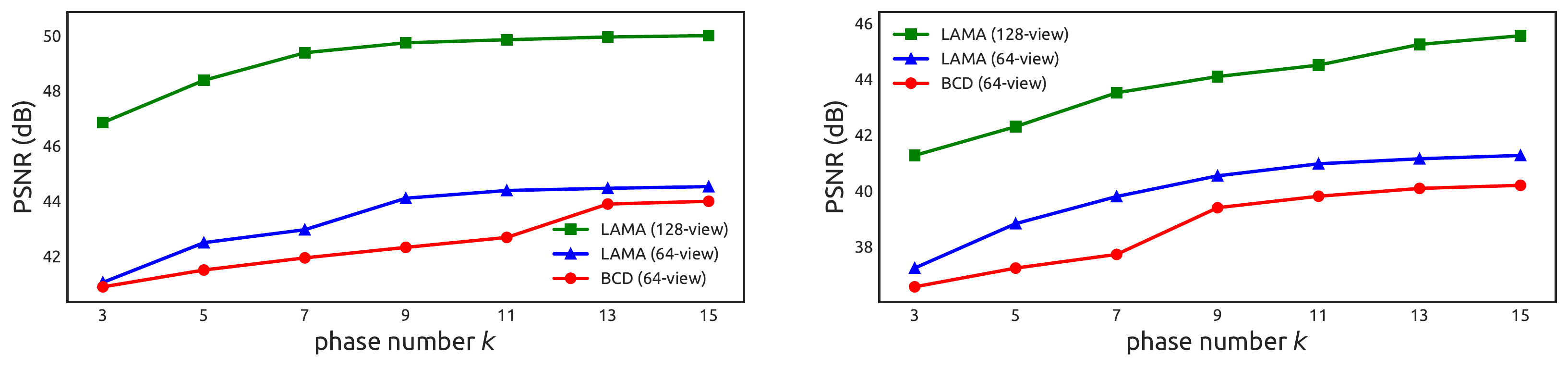}
    \caption{PSNR of reconstructions obtained by LAMA or BCD over phase number $k$ using 64-view or 128-view sinograms. \textit{Left}: AAPM-Mayo. \textit{Right}: NBIA.}
    \label{fig:ablation}
\end{figure}

We perform an ablation study to compare the reconstruction quality of LAMA and BCD defined in \eqref{v1}, \eqref{v2} versus the number of views and phases. Fig.~\ref{fig:ablation} illustrates that 15 phases strike a favorable balance between accuracy and computation.  
The residual architecture \eqref{eq:u} introduced in LAMA is also proven to be more effective than solely applying BCD for both datasets. As illustrated in Sec.~\ref{sec:convergenceproof}, the algorithm is also equipped with the added advantage of retaining convergence guarantees.

We evaluate LAMA by applying the pipeline described in Section \ref{setup} to sparse-view sinograms from the test set and compare with state-of-the-art methods where the numerical results are presented in Table \ref{tab:result}. Our method achieves superior results regarding PSNR and SSIM scores while having the second-lowest number of network parameters. The numerical results indicate the robustness and generalization ability of our approach. Additionally, we demonstrate the effectiveness of our method in preserving structural details while removing noise and artifacts through Fig.~\ref{fig:results}. More visual results are provided in the supplementary materials. Overall, our approach significantly outperforms state-of-the-art methods, as demonstrated by both numerical and visual evaluations.

\section{Conclusion}
We propose a novel, interpretable dual-domain sparse-view CT image reconstruction algorithm LAMA. It is a variational model with composite objectives and solves the nonsmooth and nonconvex optimization problem with convergence guarantees. By introducing learnable regularizations, our method effectively suppresses noise and artifacts while preserving structural details in the reconstructed images. The LAMA algorithm leverages complementary information from both domains to estimate missing information and improve reconstruction quality in each iteration. Our experiments demonstrate that LAMA outperforms existing methods while maintaining favorable memory efficiency.

\bibliographystyle{unsrt}
\bibliography{miccai}

\begin{thebibliography}{10}

\bibitem{kak_slaney_2001}
Avinash~C Kak and Malcolm Slaney.
\newblock {\em Principles of computerized tomographic imaging}.
\newblock Society For Industrial And Applied Mathematics, 2001.

\bibitem{RUDIN1992259}
Leonid~I. Rudin, Stanley Osher, and Emad Fatemi.
\newblock Nonlinear total variation based noise removal algorithms.
\newblock {\em Physica D: Nonlinear Phenomena}, 60(1):259--268, 1992.

\bibitem{accFV}
Samuel~J. LaRoque, Emil~Y. Sidky, and Xiaochuan Pan.
\newblock Accurate image reconstruction from few-view and limited-angle data in
  diffraction tomography.
\newblock {\em Journal of the Optical Society of America A}, 25(7):1772, Jun
  2008.

\bibitem{Kim_2016}
Hojin Kim, Josephine Chen, Adam Wang, Cynthia Chuang, Mareike Held, and Jean
  Pouliot.
\newblock Non-local total-variation (nltv) minimization combined with
  reweighted l1-norm for compressed sensing ct reconstruction.
\newblock {\em Physics in Medicine \& Biology}, 61(18):6878, sep 2016.

\bibitem{DDNet}
Zhicheng Zhang, Xiaokun Liang, Xu~Dong, Yaoqin Xie, and Guohua Cao.
\newblock A sparse-view ct reconstruction method based on combination of
  densenet and deconvolution.
\newblock {\em IEEE Transactions on Medical Imaging}, 37(6):1407--1417, 2018.

\bibitem{DuDoTrans}
Ce~Wang, Kun Shang, Haimiao Zhang, Qian Li, Yuan Hui, and S.~Kevin Zhou.
\newblock Dudotrans: Dual-domain transformer provides more attention for
  sinogram restoration in sparse-view ct reconstruction, 2021.

\bibitem{sinSyn}
Hoyeon Lee, Jongha Lee, Hyeongseok Kim, Byungchul Cho, and Seungryong Cho.
\newblock Deep-neural-network-based sinogram synthesis for sparse-view ct image
  reconstruction.
\newblock {\em IEEE Transactions on Radiation and Plasma Medical Sciences},
  3(2):109--119, 2019.

\bibitem{DRONE}
Weiwen Wu, Dianlin Hu, Chuang Niu, Hengyong Yu, Varut Vardhanabhuti, and
  Ge~Wang.
\newblock Drone: Dual-domain residual-based optimization network for
  sparse-view ct reconstruction.
\newblock {\em IEEE Transactions on Medical Imaging}, 40(11):3002--3014, 2021.

\bibitem{FBPConvNet}
Kyong~Hwan Jin, Michael~T. McCann, Emmanuel Froustey, and Michael Unser.
\newblock Deep convolutional neural network for inverse problems in imaging.
\newblock {\em IEEE Transactions on Image Processing}, 26(9):4509--4522, 2017.

\bibitem{LEARN}
Hu~Chen, Yi~Zhang, Yunjin Chen, Junfeng Zhang, Weihua Zhang, Huaiqiang Sun,
  Yang Lv, Peixi Liao, Jiliu Zhou, and Ge~Wang.
\newblock Learn: Learned experts’ assessment-based reconstruction network for
  sparse-data ct.
\newblock {\em IEEE Transactions on Medical Imaging}, 37(6):1333--1347, 2018.

\bibitem{GamReg}
Junfeng Zhang, Yining Hu, Jian Yang, Yang Chen, Jean-Louis Coatrieux, and Limin
  Luo.
\newblock Sparse-view x-ray ct reconstruction with gamma regularization.
\newblock {\em Neurocomputing}, 230:251--269, 2017.

\bibitem{RED-CNN}
Hu~Chen, Yi~Zhang, Mannudeep~K. Kalra, Feng Lin, Yang Chen, Peixi Liao, Jiliu
  Zhou, and Ge~Wang.
\newblock Low-dose ct with a residual encoder-decoder convolutional neural
  network.
\newblock {\em IEEE Transactions on Medical Imaging}, 36(12):2524--2535, 2017.

\bibitem{ISTANet}
Jian Zhang and Bernard Ghanem.
\newblock Ista-net: Interpretable optimization-inspired deep network for image
  compressive sensing.
\newblock In {\em 2018 IEEE/CVF Conference on Computer Vision and Pattern
  Recognition}, pages 1828--1837, 2018.

\bibitem{unrolling}
Vishal Monga, Yuelong Li, and Yonina~C. Eldar.
\newblock Algorithm unrolling: Interpretable, efficient deep learning for
  signal and image processing.
\newblock {\em IEEE Signal Processing Magazine}, 38(2):18--44, 2021.

\bibitem{learnPP}
Yi~Zhang, Hu~Chen, Wenjun Xia, Yang Chen, Baodong Liu, Yan Liu, Huaiqiang Sun,
  and Jiliu Zhou.
\newblock Learn++: Recurrent dual-domain reconstruction network for compressed
  sensing ct, 2020.

\bibitem{LDA}
Yunmei Chen, Hongcheng Liu, Xiaojing Ye, and Qingchao Zhang.
\newblock Learnable descent algorithm for nonsmooth nonconvex image
  reconstruction.
\newblock {\em SIAM Journal on Imaging Sciences}, 14(4):1532--1564, 2021.

\bibitem{TransItr}
Wenjun Xia, Ziyuan Yang, Qizheng Zhou, Zexin Lu, Zhongxian Wang, and Yi~Zhang.
\newblock A transformer-based iterative reconstruction model for sparse-view ct
  reconstruction.
\newblock In Linwei Wang, Qi~Dou, P.~Thomas Fletcher, Stefanie Speidel, and
  Shuo Li, editors, {\em Medical Image Computing and Computer Assisted
  Intervention -- MICCAI 2022}, pages 790--800, Cham, 2022. Springer Nature
  Switzerland.

\bibitem{DDPNet}
Rongjun Ge, Yuting He, Cong Xia, Hailong Sun, Yikun Zhang, Dianlin Hu, Sijie
  Chen, Yang Chen, Shuo Li, and Daoqiang Zhang.
\newblock Ddpnet: A novel dual-domain parallel network for low-dose ct
  reconstruction.
\newblock In Linwei Wang, Qi~Dou, P.~Thomas Fletcher, Stefanie Speidel, and
  Shuo Li, editors, {\em Medical Image Computing and Computer Assisted
  Intervention -- MICCAI 2022}, pages 748--757, Cham, 2022. Springer Nature
  Switzerland.

\bibitem{ldct}
Qingchao Zhang, Mehrdad Alvandipour, Wenjun Xia, Yi~Zhang, Xiaojing Ye, and
  Yunmei Chen.
\newblock Provably convergent learned inexact descent algorithm for low-dose ct
  reconstruction, 2021.

\bibitem{wanyuMiccai}
Wanyu Bian, Qingchao Zhang, Xiaojing Ye, and Yunmei Chen.
\newblock A learnable variational model for joint multimodal mri
  reconstruction and synthesis.
\newblock {\em Lecture Notes in Computer Science}, 13436:354–364, 2022.

\bibitem{ProxSARAH}
Nhan~H. Pham, Lam~M. Nguyen, Dzung~T. Phan, and Quoc Tran-Dinh.
\newblock Proxsarah: An efficient algorithmic framework for stochastic
  composite nonconvex optimization.
\newblock {\em Journal of Machine Learning Research}, 21(110):1--48, 2020.

\bibitem{PALM}
Jérôme Bolte, Shoham Sabach, and Marc Teboulle.
\newblock Proximal alternating linearized minimization for nonconvex and
  nonsmooth problems - mathematical programming, Jul 2013.

\bibitem{inertialProx}
Thomas Pock and Shoham Sabach.
\newblock Inertial proximal alternating linearized minimization (ipalm) for
  nonconvex and nonsmooth problems.
\newblock {\em SIAM Journal on Imaging Sciences}, 9(4):1756--1787, 2016.

\bibitem{stochProx}
Derek Driggs, Junqi Tang, Jingwei Liang, Mike Davies, and Carola-Bibiane
  Sch\"{o}nlieb.
\newblock A stochastic proximal alternating minimization for nonsmooth and
  nonconvex optimization.
\newblock {\em SIAM Journal on Imaging Sciences}, 14(4):1932--1970, 2021.

\bibitem{inexBlock}
Yang Yang, Marius Pesavento, Zhi-Quan Luo, and Björn Ottersten.
\newblock Inexact block coordinate descent algorithms for nonsmooth nonconvex
  optimization.
\newblock {\em IEEE Transactions on Signal Processing}, 68:947--961, 2020.

\bibitem{ResNet}
Kaiming He, Xiangyu Zhang, Shaoqing Ren, and Jian Sun.
\newblock Deep residual learning for image recognition, 2015.

\bibitem{nesterov_2004}
Yu. Nesterov.
\newblock Smooth minimization of non-smooth functions.
\newblock {\em Mathematical Programming}, 103(1):127–152, Dec 2004.

\bibitem{distDriven}
B.~De~Man and S.~Basu.
\newblock Distance-driven projection and backprojection.
\newblock In {\em 2002 IEEE Nuclear Science Symposium Conference Record},
  volume~3, pages 1477--1480 vol.3, 2002.

\bibitem{distDriven3}
Bruno~De Man and Samit Basu.
\newblock Distance-driven projection and backprojection in three dimensions.
\newblock {\em Physics in Medicine and Biology}, 49(11):2463–2475, May 2004.

\bibitem{xia2021magic}
Wenjun Xia, Zexin Lu, Yongqiang Huang, Zuoqiang Shi, Yan Liu, Hu~Chen, Yang
  Chen, Jiliu Zhou, and Yi~Zhang.
\newblock Magic: Manifold and graph integrative convolutional network for
  low-dose ct reconstruction.
\newblock {\em IEEE Transactions on Medical Imaging}, 2021.

\end{thebibliography}
\newpage
\section*{Supplementary Materials}
\begin{figure}
    \centering
    \includegraphics[scale=0.59]{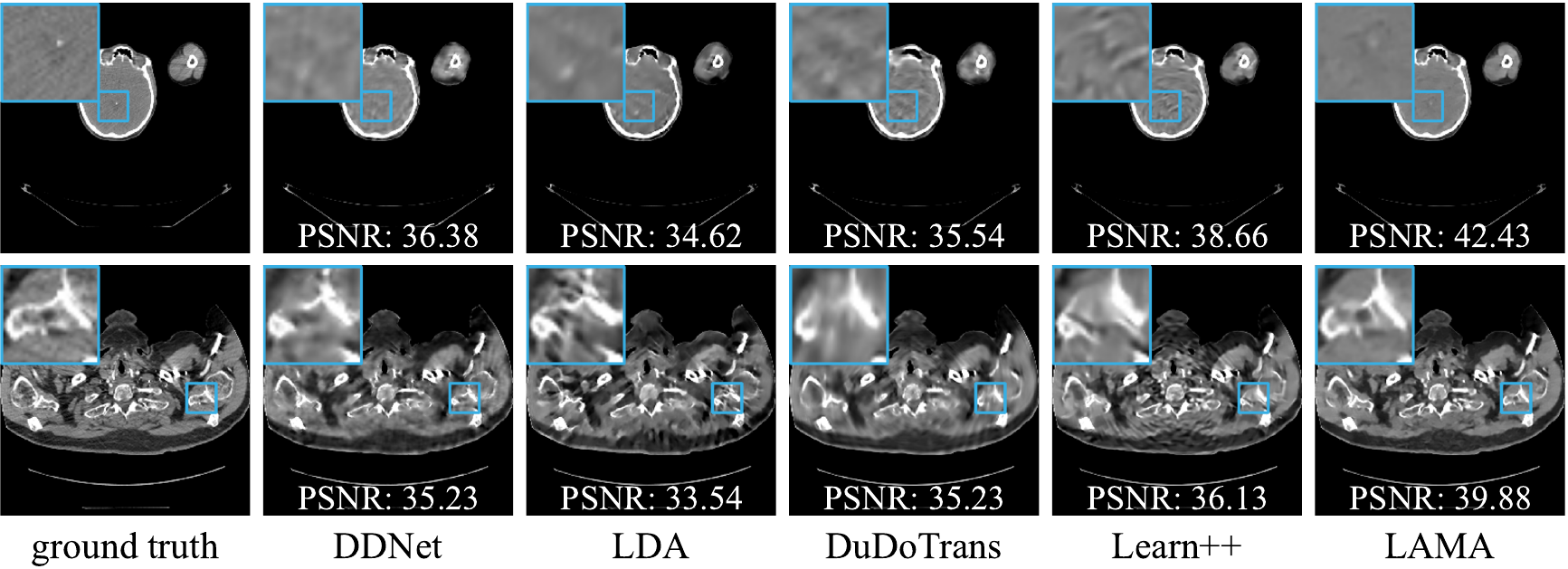}
    \caption{Visual comparison for NBIA dataset using 64-view sinograms.}
    \label{fig:NBIAComparison}
\end{figure}
Next, we give a brief proof for Theorem 5.01. Let's start with a few lemmas.
For a function $f(\xbf, \zbf)$ the vector $(\nabla_\xbf f, \nabla_\zbf f)$ is denoted by $\nabla_{\xbf,\zbf} f $.

\begin{lemma}\label{lem:r_subdiff}
The Clarke subdifferential of $\|g^r(\ybf)\|_{2,1}$, where $\*g^r(\cdot)\in\mathbb{R}^{m_r\times d_r}$ and $r$ is either $R$ or $Q$ defined in \eqref{eq:21norm}, is as follows:
\begin{align}\label{eq:r_subdiff}
\partial^c \|g^r(\ybf)\|_{2,1} = \bigg\{&\sum_{i\in I_0}\nabla g^r_i(\ybf)^{\top}  \wbf_i + \sum_{i \in I_1}\nabla g^r_i(\ybf)^{\top}\frac{g^r_i(\ybf)}{\|g^r_i(\ybf)\|} \ \bigg\vert \nonumber \\
&\ \wbf_i \in \mathbb{R}^{d_r}, \ \|\Pi(\wbf_i; \Ccal(\nabla g^r_i(\ybf)))\|\leq 1,\ \forall\, i \in I_0 \bigg\} ,  
\end{align}
where $I_0=\{i \in [m_r] \ | \ \|g^r_i(\ybf) \|= 0 \}$, $I_1=[m_r] \setminus I_0$, and $\Pi(\wbf;\Ccal(\Abf))$ is the projection of $\wbf$ onto $\Ccal(\Abf)$ which stands for the column space of $\Abf$. 
\end{lemma}

\begin{lemma}
    The gradient
    $\nabla r_{\varepsilon}(\ybf)$ in \eqref{eq:smoothedR} is ($\sqrt{m} L_{g^r}+\frac{M^2}{\varepsilon}$)-Lipschitz continuous, 
    where $L_{g^r}$ is the Lipschitz constant of $\nabla g^r$ and $M=\sup_{y} \|\nabla g^r(y) \|_2$. Consequently, $\nabla_{\xbf,\zbf} \Phi_\varepsilon(\xbf, \zbf)$ is $L_{\varepsilon}$-Lipschitz continuous with $L_{\varepsilon}=O(\varepsilon^{-1})$.

\end{lemma}

The proofs of Lemmas 7.01-7.02 can be found from Lemmas 3.1-3.2 in \cite{LDA}.

\begin{lemma}
    Let $\Phi_\varepsilon(\xbf, \zbf)$ be 
    defined in \eqref{SM}, and $(\xbf_0, \zbf_0) \in X \times Z$ be arbitrary. Suppose $\{\xbf_{k}, \zbf_{k}\}$ is the sequence generated by repeating Lines 2--10 of Algorithm 1  with $\varepsilon_k = \varepsilon$. Then $\norm{\nabla\Phi_{\varepsilon}(\xbf_k,\zbf_k)}\to 0$ as $k\to\infty$.

\end{lemma}

\noindent{\bf Proof:} 
Given $(\xbf_k, \zbf_k)$, in case that 
$(\xbf_{k+1},\zbf_{k+1})=(\*v^{\*x}_{k+1},\*v^{\*z}_{k+1})$, and condition (13) holds after backtracking for $\ell_k \geq 0$ times. 
Then the algorithm computes
\begin{equation}\label{v-12c}
\*v^{\*x}_{k+1}=\xbf_k-\bar \alpha \rho^{\ell_k}\nabla_{\xbf}\Phi_{\varepsilon}(\xbf_k,\zbf_k), \qquad \*v^{\*z}_{k+1}=\zbf_k-\bar \beta \rho^{\ell_k}\nabla_{\zbf}\Phi_{\varepsilon}(\*v^{\*x}_{k+1},\zbf_k).
\end{equation}

By the $L_{\varepsilon}$-Lipschitz continuity of $\nabla_{\xbf,\zbf} \Phi_{\varepsilon}$ and \eqref{v-12c}, we have 
\begin{align}\label{phi-to-v}
&\Phi_{\varepsilon}(\*v^{\*x}_{k+1},\*v^{\*z}_{k+1})
\le \Phi_{\varepsilon}(\*v^{\*x}_{k+1},\zbf_k)+\nabla_{\zbf}\Phi_{\varepsilon}(\*v^{\*x}_{k+1},\zbf_k)\cdot
(\*v^{\*z}_{k+1}-\zbf_k)+\frac{L_{\varepsilon}}{2}\|\*v^{\*z}_{k+1}-\zbf_k\|^2\nonumber\\
& \ \le \Phi_{\varepsilon}(\xbf_k,\zbf_k)+\nabla_x\Phi_{\varepsilon}(\xbf_k,\zbf_k)\cdot (\*v^{\*x}_{k+1}-\xbf_k)+\frac{L_{\varepsilon}}{2}\|\*v^{\*x}_{k+1}-\xbf_k\|^2\nonumber\\
& \quad \ +\nabla_z\Phi_{\varepsilon}(\*v^{\*x}_{k+1},\zbf_k)\cdot
(\*v^{\*z}_{k+1}-\zbf_k)+\frac{L_{\varepsilon}}{2}\|\*v^{\*z}_{k+1}-\zbf_k\|^2 \\
& \leq   \Phi_{\varepsilon}(\xbf_k,\zbf_k)+(-\frac 1{\bar \alpha\rho^{l_k}}+\frac{L_{\epsilon}}{2})\|\*v^{\*x}_{k+1}-\xbf_k\|^2
+(-\frac{1}{\bar \beta\rho^{l_k}}+\frac{L_{\varepsilon}}{2})\|\*v^{\*z}_{k+1}-\zbf_k\|^2. \nonumber
\end{align}
Hence, for any $k=1,2,\ldots$ the maximum line search steps $\ell_{max}$ required for \eqref{v-condition-6} satisfies $\rho^{\ell_{max}}= (\delta+L_{\varepsilon}/2)^{-1}(\max\{\bar \alpha,\bar \beta\})^{-1}$. Note that  
$0 \le \ell_k \le  \ell_{max}$, hence,
\begin{equation} \label{rho-to-lmax}
  \min\{\bar \alpha,\bar \beta\}  \rho^{\ell_{k}}\geq 
  \min\{\bar \alpha,\bar \beta\} (\delta+L_{\varepsilon}/2)^{-1}(\max\{\bar \alpha,\bar \beta\})^{-1}.
    \end{equation}
Moreover, from \eqref{v-12c} 
we know that when the condition \eqref{v-condition-6} is met, 
we have
\begin{align} \label{smoothphi-decay2}
& \Phi_{\varepsilon}(\*v^{\*x}_{k+1},\*v^{\*z}_{k+1})-   \Phi_{\varepsilon}(\xbf_k,\zbf_k) \nonumber \\ 
& \leq -\delta (\bar \alpha\rho^{\ell_k})^2\|\nabla_x\Phi_{\varepsilon}(\xbf_k,\zbf_k)\| ^2-\delta (\bar \beta \rho^{\ell_k})^2\|\nabla_z\Phi_{\varepsilon}(\*v^{\*x}_{k+1},\zbf_k)\|^2.
\end{align}

Now we estimate the last term in \eqref{smoothphi-decay2}.
Using $L_{\varepsilon}$-Lipschitz of $\nabla_{x,z} \Phi_{\varepsilon}$ and the inequality 
$b^2\ge \mu a^2-\frac{\mu}{1-\mu}(a- b)^2$ for all $\mu\in(0,1)$,  we  get
\begin{align} \label{tem-1}
  & \|\nabla_z\Phi_{\varepsilon}(\*v^{\*x}_{k+1},\zbf_k)\|^2 - \mu\|\nabla_z\Phi_{\varepsilon}(\xbf_{k},\zbf_k)\|^2 \geq -\frac{\mu}{1-\mu}\|\nabla_z\Phi_{\varepsilon}(\*v^{\*x}_{k+1},\zbf_k)-\nabla_z\Phi_{\varepsilon}(\xbf_{k},\zbf_k)\|^2 \nonumber \\   
 & \geq 
 -\frac{\mu}{1-\mu}
  L_{\varepsilon}^2\|\*v^{\*x}_{k+1}-\xbf_k\|^2 
=-\frac{\mu}{1-\mu}
  L_{\varepsilon}^2 (\bar \alpha \rho^{\ell_k})^2\|\nabla_x\Phi_{\varepsilon}(\xbf_{k},\zbf_k)\|^2.
\end{align}

Inserting \eqref{tem-1} into \eqref{smoothphi-decay2}, we have 
\begin{align} \label{line-up-3}
&\Phi_{\varepsilon}(\*v^{\*x}_{k+1},\*v^{\*z}_{k+1})-\Phi_{\varepsilon}(\xbf_k,\zbf_k)
  \le -\delta\mu(\bar \beta \rho^{\ell_k})^2\|\nabla_z\Phi_{\varepsilon}(\xbf_{k},\zbf_k)\|^2\nonumber\\
  & -\delta
  (\bar \alpha \rho^{\ell_k})^2\big(1-\frac{\mu}{1-\mu}(\bar \beta \rho^{\ell_k})^2 L_{\varepsilon}^2\big )\|\nabla_x\Phi_{\varepsilon}(\xbf_{k},\zbf_k)\|^2. 
\end{align}
Then from \eqref{rho-to-lmax} and \eqref{line-up-3}, there are a sufficiently small $\mu$ and a constant $C_1>0$, depending only on $\rho$, $\delta$, $\bar \alpha$, $\bar \beta$, such that 
\begin{equation}
 \|\nabla_{x,z}\Phi_{\varepsilon}(\xbf_{k},\zbf_k)\|^2   \leq C_1L_{\varepsilon}^2\big(\Phi_{\varepsilon}(\xbf_k,\zbf_k)-\Phi_{\varepsilon}(\*v^{\*x}_{k+1},\*v^{\*z}_{k+1})\big).
\end{equation}

On the other hand, in case that we can take $(\xbf_{k+1},\zbf_{k+1})=(\*u^{\*x}_{k+1},\*u^{\*z}_{k+1})$, then  the conditions in (10a-b) hold. 
Hence in any case we have
    \begin{align}
    \label{eq:xconda}
    \Phi_{\varepsilon}({\*x}_{k+1},{\*z}_{k+1})-\Phi_{\varepsilon}(\*x_k,\*z_k) &\leq -C_2\left(\norm{{\*x}_{k+1}-\*x_k}^2+\norm{{\*z}_{k+1}-\*z_k}^2\right),\\
    \label{eq:xcondb}
       \norm{\nabla \Phi_{\varepsilon}(\*x_{k},\*z_{k})}^2&\leq C_3\big(\Phi_{\varepsilon}(\xbf_k,\zbf_k)-\Phi_{\varepsilon}(\*v^{\*x}_{k+1},\*v^{\*z}_{k+1})\big),
    \end{align}
    where $C_2=\min\{\eta, \delta \}$, $C_3= \max\{2/\eta^3, C_1L_{\varepsilon}^2\}$ 
    From \eqref{eq:xconda}, It is easy to conclude that there is a $\Phi_{\varepsilon}^*$, such that $\nabla \Phi_{\varepsilon}(\*x_{k},\*z_{k})\downarrow \Phi_{\varepsilon}^*$ as $k\to\infty$. Then adding up both sides of  \eqref{eq:xcondb} w.r.t. $k$, it yields that for any $K$, $\sum_{k=0}^K \norm{\nabla \Phi_{\varepsilon}(\*x_{k},\*z_{k})}^2\leq C_3(\Phi_{\varepsilon}(\*x_0,\*z_0)-\Phi_{\varepsilon}^*)$.  This leads to the conclusion of the lemma immediately.

   {\bf Proof of Theorem 5.01:} By using the above three lemmas the theorem can be proved by the argument similar to Theorem 3.6 in \cite{LDA}.

\end{document}